\documentclass[11pt,a4paper]{article}
\usepackage[T1]{fontenc}
\usepackage{mathptmx}
\usepackage[margin=2.3cm]{geometry}
\usepackage{cite}
\usepackage{amsmath,amssymb}
\usepackage{algorithm}
\usepackage{algorithmic}
\usepackage{graphicx}
\usepackage{textcomp}
\usepackage{booktabs}
\usepackage{bm}
\usepackage[hidelinks]{hyperref}
\usepackage{authblk}

\usepackage{fancyhdr}
\fancypagestyle{plain}{\fancyhf{}\fancyhead[C]{\footnotesize\itshape Tree Depth and Sample-Quality-Associated Features: Two Hidden Factors in Boruta Selection for RNA-Seq Data Classification}\fancyfoot[L]{\small Chowdhury \textit{et al.}}\fancyfoot[R]{\small\thepage}}

\title{Tree Depth and Sample-Quality-Associated Features: Two Hidden Factors in Boruta Selection for RNA-Seq Data Classification}
\author[1]{Tareque Mohmud Chowdhury\thanks{Corresponding author: tareque@iut-dhaka.edu}}
\author[1]{Abu Raihan Mostofa Kamal}
\affil[1]{Department of Computer Science and Engineering, Islamic University of Technology, Gazipur 1704, Bangladesh}
\date{}

\begin{document}
\maketitle

\begin{abstract}
Random Forest--based selectors such as Boruta depend on tree depth, yet Boruta's most used implementation, BorutaPy, recommends depths of 3 to 7 without published evidence. We tested the effect of depth using pan-cancer data from The Cancer Genome Atlas (10,469 tumors, 33 cancer types) and breast cancer PAM50 subtyping (981 tumors, 5 subtypes). Label-dependent steps were fitted inside training folds, effects were measured against seed-to-seed variation, and each task kept a locked test set. In pan-cancer data, depth 5 changed the selection far more than a new seed did: the top-500 Jaccard overlap with unlimited depth was 0.37, against 0.84 (depth 5) and 0.93 (unlimited) between seeds. Depth-5 trees confirmed 472 instead of 1,659 genes, covered fewer large-class markers (16\% vs 62\%) and gave less accurate 50-gene panels (balanced accuracy 0.809 vs 0.889). The effect vanished once the cap reached 61 to 76\% of the median depth of unrestricted trees, a depth that depends mainly on the number of classes, not sample size. In PAM50 subtyping (median depth 12) the effect was small, and leaving depth uncapped cost no accuracy in either task. It replicated in a second cross-validation repeat and under permutation importance. About a third of pre-filtered candidates correlated within class with the number of detected genes, a sample-quality measure; a within-class filter removed them without loss of test accuracy (0.943 vs 0.942 balanced accuracy, 1,000 genes). We recommend uncapped trees, or a cap above 75\% of the realized median depth, and a within-class quality filter before Boruta.

\medskip\noindent\textbf{Keywords:}	Cancer classification, feature selection, gene expression, random forest, boruta, hyperparameter sensitivity, machine learning, RNA sequencing, The Cancer Genome Atlas.
\end{abstract}

\section{Introduction}
\label{sec:introduction}
RNA sequencing measures the expression of about 60,000 genes in each sample, while a single cancer cohort usually contains only hundreds to a few thousand samples \cite{wang2009,weinstein2013}. Classifiers of cancer type and subtype built on such data depend on feature selection: a small gene panel is cheaper to measure, easier to interpret and easier to validate in new cohorts than a model that uses every gene \cite{guyon2003,saeys2007,alhenawi2022}. The 50-gene PAM50 assay for breast cancer subtypes is a well-known example of such a panel \cite{parker2009}. For a panel to be useful, the selection method must find genes that carry real class signal, and it must find them reproducibly. Classifiers of cancer type trained on The Cancer Genome Atlas (TCGA) reach high accuracy with both classical and deep learning models \cite{li2017,khorshed2020,mostavi2020,mohammed2021}. These studies report how accurate the final classifier is, but not how the selected genes depend on the settings of the selection method.

Boruta is a popular selection method for this task \cite{kursa2010}. It adds a permuted copy, or shadow, of every feature, fits a Random Forest \cite{breiman2001}, and keeps the features whose importance repeatedly exceeds that of the best shadow. Because it seeks all relevant features rather than a panel of fixed size, it suits exploratory biology, and it has been found to be among the most self-consistent Random Forest selectors on gene expression data \cite{diazuriarte2006,kursa2014,degenhardt2019}. Its most used Python implementation, BorutaPy, recommends ``pruned trees with a depth between 3-7'' and shows a maximum depth of 5 in its example \cite{borutapy}. To our knowledge, no published study supports this advice. In large multi-class problems, unrestricted trees grow far deeper: in our pan-cancer data their median depth was 33. A shallow cap therefore removes most of each tree's splits, and impurity importance can only credit features that are used in the splits that remain.

This mismatch leads to the first of two research gaps that this study addresses: how Random Forest tree depth changes the genes that Boruta selects. The second is whether candidate genes track sample quality rather than biology. On the first gap, studies of Random Forest parameters have tuned the number of candidate features per split, the sample fraction and the minimal node size, mostly for prediction accuracy or in small problems, where trees stay shallow whatever the setting \cite{wright2016,probst2019}. To our knowledge, the effect of tree depth on feature selection in large multi-class data has not been measured. On the second gap, candidate genes can track technical differences between samples rather than biology. In TCGA RNA sequencing (RNA-seq) data, library size, tumor purity and processing plate leave signal that normalization does not remove \cite{leek2010,aran2015,molania2023}, and in single-cell data the number of detected genes per cell, the quality measure we use here, drives a large share of variation \cite{hicks2018}. Selectors that reserve places for non-coding RNA categories may draw such genes in, because a quota must be filled even when the genes available to fill it carry little class signal. Removing whole gene types is not a solution, because pseudogene and long non-coding RNA (lncRNA) expression can carry real subtype signal \cite{kalyanasundaram2012,han2014,yan2015}. To our knowledge, no feature selection study has tested candidate genes for within-class association with sample quality before selection.

Measuring either effect reliably requires an evaluation design that avoids two pitfalls. Selecting features outside the cross-validation loop inflates accuracy \cite{ambroise2002,varma2006}, and an apparent difference between two settings may be no larger than the difference between two random seeds of the same setting \cite{kalousis2007,nogueira2018}. We therefore used a protocol in which every label-dependent step is fitted inside training folds, each task has a locked test set opened once, and every comparison of settings is measured against the variation between seeds.

This paper makes four contributions:
\begin{enumerate}
\item To our knowledge, the first systematic study of Random Forest depth inside Boruta on large multi-class data: up to nine depth settings in two tasks (33-class pan-cancer and 5-class PAM50 subtyping), with effects measured against seed noise and replicated across cross-validation repeats and importance measures.
\item Evidence that the size of the effect follows the depth that unrestricted trees reach, which is set mainly by the number of classes rather than by sample size, and a practical rule derived from it: leave depth uncapped, which cost no accuracy or stability in either task, or check the realized depth before choosing a cap.
\item A within-class quality-association measure showing that about a third of pre-filtered candidate genes track sample quality, and a filter that removes them before selection without loss of test accuracy.
\item A leakage-free, reproducible selection pipeline, evaluated on locked test sets, with code that reproduces every analysis in this paper.
\end{enumerate}

Section~\ref{sec:related} reviews related work. Section~\ref{sec:methods} describes the data, pipeline and experiments, Section~\ref{sec:results} the results, and Section~\ref{sec:discussion} their implications and limitations. Section~\ref{sec:conclusion} concludes.

\section{Related Work}
\label{sec:related}

\subsection{Feature Selection for Gene Expression Classification}
Feature selection for gene expression data falls into filter methods, which rank genes by a univariate statistic; wrapper methods, which search gene subsets with a classifier; and embedded methods, which select during model fitting \cite{guyon2003,saeys2007,boloncanedo2014}. Hybrid designs combine a fast filter with a slower wrapper or embedded step \cite{alhenawi2022}. Common first-stage filters include univariate tests and minimum-redundancy maximum-relevance ranking \cite{peng2005}, and embedded selection with boosted models has been used to build cancer gene signatures \cite{seo2020}. Boruta can serve as the second stage of such designs, as in our pipeline. Studies of these hybrids report the accuracy of the final panel; to our knowledge, none has examined how Boruta's tree depth changes the genes selected.

\subsection{Random Forest Importance and Its Parameters}
Boruta relies on Random Forest importance, by default the mean decrease in impurity, which accumulates over the splits in which a feature is used \cite{breiman2001,louppe2013}. Impurity importance can favor features with many possible split points, and permutation importance is the usual alternative \cite{strobl2007,altmann2010,nembrini2018}. Wright \emph{et al.} studied how the number of candidate features per split affects the detection of interacting variables, with trees grown to purity \cite{wright2016}. Janitza \emph{et al.} proposed Vita, a fast importance test for high-dimensional data, as an alternative to Boruta \cite{janitza2018}. Kursa compared Random Forest selectors on microarray data and found Boruta the most self-consistent; deeper Random Ferns confirmed more genes, but Random Forest depth was not varied \cite{kursa2014}. Degenhardt \emph{et al.} compared Random Forest selection methods on omics data at fixed settings \cite{degenhardt2019}. Reviews of Random Forest tuning discuss tree size, through node size or depth, as a parameter for prediction accuracy rather than for feature selection \cite{probst2019}. To our knowledge, no peer-reviewed study has varied Random Forest depth inside Boruta on large multi-class data or tested the depth range recommended by BorutaPy.

\subsection{Stability of Feature Selection}
A selection method should return similar genes when the data change slightly. The Nogueira index measures this across resamples and corrects for set size \cite{nogueira2018}; an earlier index is that of Kalousis \emph{et al.} \cite{kalousis2007}. Gene signatures for the same clinical question are known to overlap little between studies \cite{eindor2005}. Most stability studies compare methods; few separate the variation caused by a setting from the variation between random seeds of the same setting, which is the comparison our design uses.

\subsection{Unwanted Variation in RNA-Seq Data}
Molania \emph{et al.} showed that TCGA RNA-seq data carry variation from library size, tumor purity and processing plates that common normalizations do not remove, and that it distorts downstream analyses, including subtype calls \cite{molania2023}. Batch effects are common in high-throughput data \cite{leek2010}, tumor purity varies widely across TCGA samples \cite{aran2015}, and RNA degradation changes transcript quantification \cite{gallegoromero2014}. In single-cell data, the fraction of detected genes per cell drives a major axis of variation \cite{hicks2018}. Pseudogene expression, often treated as noise, carries subtype-specific signal across cancers \cite{kalyanasundaram2012,han2014}, and lncRNAs show cancer-type-specific expression \cite{iyer2015,yan2015}. Our quality-association measure applies these observations to feature selection: it tests each candidate gene, within class, for association with the number of detected genes per sample, instead of correcting the whole matrix or excluding gene types.

\section{Materials and Methods}
\label{sec:methods}
\begin{figure*}[!t]
\centering
\includegraphics[width=\textwidth]{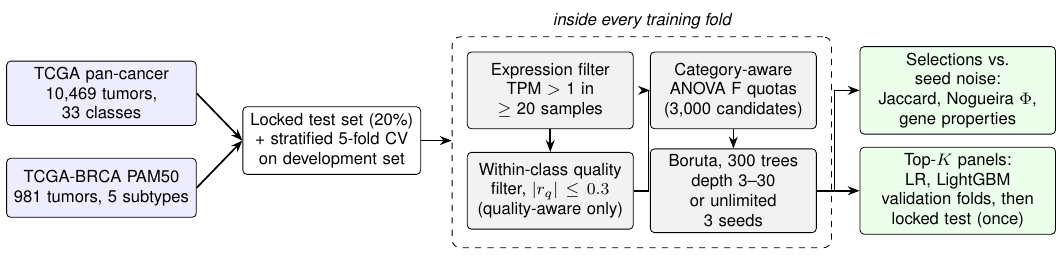}
\caption{Study design. Two tasks from TCGA RNA-seq data: pan-cancer classification (10,469 tumors, 33 classes) and breast cancer PAM50 subtyping (981 tumors, 5 classes). Each task has a locked test set (20\% of patients). Inside every training fold: expression filter, within-class quality filter (quality-aware pipeline only), category-aware ANOVA F pre-filter to 3,000 candidates, and Boruta at each depth setting with three random seeds. Selections are compared with seed-to-seed variation; panels of the top $K$ genes are evaluated with logistic regression (LR) and LightGBM.}\label{fig:design}
\end{figure*}

All experiments followed two rules: every step that uses class labels was fitted on training samples only, and each task's test set was opened once, after all choices were fixed. Fig.~\ref{fig:design} summarizes the design.

\subsection{Data}
RNA-seq gene expression data (STAR--Counts workflow \cite{dobin2013}, GENCODE v36 annotation \cite{frankish2021}) for all 33 TCGA projects \cite{weinstein2013} were downloaded from the NCI Genomic Data Commons (GDC) \cite{grossman2016} through its REST API (Data Release 46.0, August 10, 2026; downloaded October 15--16, 2026). In total, 11,505 open-access files with 60,660 genes each were retrieved and checked against their GDC MD5 checksums. Expression was represented as $\log_2(\mathrm{TPM}+1)$, where TPM denotes transcripts per million \cite{li2011}. Plate and tissue source site identifiers were parsed from aliquot barcodes for the batch check described in Section~\ref{sec:cohort}.

Genes were assigned to four RNA categories from their GENCODE gene types: protein-coding (mRNA, 19,962 genes), lncRNA (16,901), miRNA genes (1,881) and all other types (``other'', 21,916; mostly processed pseudogenes, 10,167). TCGA RNA-seq measures miRNA precursor and host transcripts, not mature miRNAs.

\subsection{Cohort, Quality Control and Splits}
\label{sec:cohort}
Primary tumor samples (sample type 01) were kept, with two exceptions: sample type 03 for LAML, which has no type 01 samples, and both primary and metastatic (06) samples for SKCM, because most SKCM tumors in TCGA are metastases. One sample per patient was chosen by a fixed rule (primary before metastatic, then vial, portion and barcode order).

For each sample we computed library size and the number of detected genes (TPM $>$ 1). Samples whose number of detected genes was more than three median absolute deviations \cite{rousseeuw1993} below their cancer-type median and below 70\% of that median were excluded (19 samples, at most 3 per cancer type). A further 253 samples, flagged by a looser rule (library size or detected genes more than three median absolute deviations below the median), were kept. The rules use quality metrics only, not class labels. In the eight largest projects, plate and site explained 2--3 times more variance in the first principal components than expected by chance (up to 4.6 times in UCEC). Because these factors are confounded with cancer type and patient population, correcting for them would also remove class signal \cite{leek2010}, so no batch correction was applied.

The final pan-cancer cohort had 10,469 tumors from 33 cancer types (Table~\ref{tab:cohorts}). Splits were stratified by cancer type, with esophageal carcinoma further split by recorded histology into adenocarcinoma and squamous cell carcinoma; a histology group with fewer than 13 samples joined the largest group of its cancer type. Twenty percent of patients (2,095) formed a locked test set. The other 8,374 were assigned to three repeats of stratified 5-fold cross-validation (CV) with the same strata (seed 42). Repeat 0 was used for all development experiments and repeat 1 for the replication check (Section~\ref{sec:robust-methods}). The smallest class in any validation fold had 5 samples (CHOL). Classifiers used the 33 cancer types as classes; the histology split served only to balance the splits.

\begin{table}[!t]
\caption{Cohorts and Splits}
\label{tab:cohorts}
\centering
\setlength{\tabcolsep}{4pt}
\begin{tabular}{lcc}
\toprule
 & Pan-cancer & PAM50 \\
\midrule
Classes & 33 cancer types & 5 subtypes \\
Tumors after QC & 10,469 & 981 \\
Development set & 8,374 & 797 \\
Locked test set & 2,095 & 184 \\
Training samples per CV fold & about 6,700 & 637 \\
Smallest class & CHOL (35) & Normal-like (36) \\
Largest class & BRCA (1,095) & LumA (499) \\
\bottomrule
\multicolumn{3}{p{0.92\columnwidth}}{\footnotesize PAM50 development subtypes: LumA 419, LumB 155, Basal 129, Her2 65, Normal-like 29; test: 80, 42, 42, 13, 7. QC: quality control.}
\end{tabular}
\end{table}

\subsection{Second Task: Breast Cancer PAM50 Subtypes}
PAM50 subtype labels \cite{parker2009} from the TCGA PanCancer Atlas \cite{hoadley2018} were obtained from cBioPortal \cite{cerami2012,gao2013} (study brca\_tcga\_pan\_can\_atlas\_2018) for 981 patients and matched to TCGA-BRCA tumors \cite{tcga2012} of the pan-cancer cohort. Samples kept their pan-cancer set: 797 development samples and 184 locked test samples (Table~\ref{tab:cohorts}). Development samples were split by stratified 5-fold CV (seed 42; training folds of 637). The design, depth grid and measures for this task were fixed before any result was seen.

\subsection{Candidate Pre-Filter}
Within each training fold, genes were first filtered without labels (TPM $>$ 1 in at least 20 training samples), which left about 48,000 genes. A category-aware analysis of variance (ANOVA) F filter then kept 3,000 candidates, with quotas proportional to each RNA category's share of the expressed genes (at least 50 per category), so that non-coding genes were not crowded out by mRNAs. In the quality-aware pipeline, the quality filter described next is applied between these two steps.

\subsection{Quality Association and Quality Filter}
\label{sec:quality-methods}
For each expressed gene, the quality association $r_q$ was the Pearson correlation, within class, between its expression and $\log_{10}$ of the sample's number of detected genes. Both variables were centered on their class means in the training fold before the correlation was computed, so differences in quality between classes do not count. A gene was quality-associated when $|r_q| > 0.3$. The quality-aware pipeline drops these genes after the expression filter and before the category quotas; thresholds 0.2 and 0.4 were run as a sensitivity check. We call the same pipeline without this filter the baseline pipeline; the depth experiment used it, so that depth effects were measured before any quality filtering. The 0.3 threshold was chosen during exploration from quality association alone, not from accuracy. Two outcomes were fixed in advance: the filter should lower the quality-associated share of the selected genes, and it should narrow the accuracy gap between high- and low-quality validation samples.

\subsection{Boruta}
Boruta \cite{kursa2010} was implemented directly, following its Python implementation BorutaPy \cite{borutapy}: each iteration adds a permuted copy of every active feature (shadow features), fits a Random Forest, and counts a hit for each gene whose impurity importance exceeds the largest shadow importance. Genes are confirmed or rejected by a two-step correction (Benjamini--Hochberg \cite{benjamini1995}, then Bonferroni over iterations) at $\alpha = 0.05$. We used 300 trees at every depth (BorutaPy's automatic tree count depends on depth, which would confound the comparison), balanced class weights, and at most 50 iterations. Each run recorded every gene's decision, hit rate (the share of iterations in which it beat the best shadow) and mean importance; genes still tentative at the end counted as not confirmed. Genes were ranked by their hit rate averaged over the three seeds, with mean importance breaking ties. Algorithm~\ref{alg:pipeline} gives the full pipeline for one training fold.

\begin{algorithm}[!t]
\caption{Quality-Aware Hybrid Selection for One Training Fold}
\label{alg:pipeline}
\begin{algorithmic}[1]
\REQUIRE training samples $X$, labels $y$, detected-gene count $q$ per sample; depth $d$, seeds $S$, candidate count $C = 3{,}000$, threshold $\tau = 0.3$
\STATE Expression filter: keep genes with TPM $>$ 1 in at least 20 training samples
\STATE Quality filter: for each gene, compute $r_q$, its within-class correlation with $\log_{10} q$; drop genes with $|r_q| > \tau$
\STATE Category-aware pre-filter: rank the remaining genes by ANOVA F; keep $C$ genes, split across RNA categories in proportion to their sizes (at least 50 each)
\FOR{each seed $s \in S$}
\STATE Run Boruta (300 trees, maximum depth $d$, at most 50 iterations, $\alpha = 0.05$); record each gene's decision, hit rate and mean importance
\ENDFOR
\STATE Rank genes by mean hit rate, then mean importance, across seeds
\ENSURE the top-$K$ genes for any panel size $K$
\end{algorithmic}
\end{algorithm}

\subsection{Depth Experiment}
Random Forest maximum depth was set to 3, 5, 8, 10, 15, 20, 25, 30 or unlimited (pan-cancer) and 3, 5, 8, 10, 15, 20 or unlimited (PAM50). The pan-cancer experiment ran in two phases. The first phase, which tested the pre-declared decision rule below, used depths 5, 15 and unlimited with seeds 0--4; the second phase added the other six depths with seeds 0--2. All comparisons across the full depth grid (Table~\ref{tab:props}, Figs.~\ref{fig:depth} and~\ref{fig:gap}) use seeds 0--2 at every depth, so every depth is represented equally. The PAM50 experiment used seeds 0--2 at every depth. Agreement between selections was measured by the Jaccard index \cite{jaccard1912} of the top-$K$ genes ($K = 500$ pan-cancer; 200 and 500 PAM50). The yardstick was seed noise: overlap between two seeds at the same depth. The depth effect was measured against unlimited depth with different seeds (seed $a$ at depth $d$ vs seed $b$ at unlimited depth, $a \neq b$), because runs sharing a seed share shadow permutations and tree random states, which inflates their overlap when the cap rarely binds. We call the difference between seed-to-seed overlap and this cross-seed overlap with unlimited depth the \emph{depth gap}. Stability across folds was measured by the Nogueira index $\Phi$ of the top-$K$ sets \cite{nogueira2018}. In the first phase we also tested whether combining depths stabilizes the selection. At equal cost (three Boruta runs per fold), genes were ranked either by hit rate averaged over depths 5, 15 and unlimited with one seed, or by hit rate averaged over three seeds at a single depth; $\Phi$ was averaged over five choices of seeds.

The decision rule for the pan-cancer task was set before the first phase and evaluated on its runs: the depth effect is real if the mean top-500 Jaccard between depths (pairs of depths 5, 15 and unlimited run with the same seed) is at least 0.05 below the mean Jaccard between seeds at the same depth, and lower in all five folds. Depth pairs that share a seed also share shadow permutations, which can only raise their overlap, so this rule is conservative.

\subsection{Gene Properties}
For each top-500 set we recorded the share of each RNA category, the median ANOVA F percentile among candidates, and marker coverage: the share of each class's 20 strongest one-vs-rest markers (training fold) found in the set, averaged over the eight smallest and the eight largest classes. For PAM50 we counted the PAM50 genes recovered in the top 100 and 200.

\subsection{Classifiers and Accuracy}
Gene sets were the top $K$ genes of each ranking (pan-cancer $K = 50$, 100, 200, 500, 1,000; PAM50 $K = 20$, 50, 100, 200, 500). Two classifiers were trained with fixed settings for every gene set: logistic regression \cite{pedregosa2011} (standardized features, L2 penalty, $C = 0.1$, balanced class weights) and LightGBM 4.7.0 \cite{ke2017} (300 trees, learning rate 0.05, 31 leaves, balanced class weights, deterministic mode). Accuracy was measured as balanced accuracy \cite{brodersen2010} on validation folds. Settings were compared by paired differences across folds. As references, we also ranked the candidates by ANOVA F and, in the pan-cancer task, drew random genes from the candidates.

\subsection{Locked Test Sets}
The final pipelines were rerun on all development samples (unlimited depth unless stated, three seeds each). Panel size $K$ was chosen per classifier by the one-standard-error rule \cite{hastie2009} on CV results of the quality-aware pipeline: the smallest $K$ whose mean balanced accuracy was within one standard error of the best. A compact panel was also reported ($K = 100$ pan-cancer; $K = 50$ PAM50, to match the PAM50 gene count). On the test sets we compared the quality-aware and baseline pipelines in both tasks, a depth-5 pipeline in PAM50, and the 50 PAM50 genes as an upper reference, since the PAM50 labels were derived from those genes. We report balanced accuracy with 95\% confidence intervals from 2,000 bootstrap resamples \cite{efron1993} drawn within class (Table~\ref{tab:test}, Fig.~\ref{fig:test}) and, where it matters, per-class recall (Section~\ref{sec:res-test}). Paired differences were tested by the bootstrap interval of the difference and an exact McNemar test \cite{mcnemar1947,dietterich1998}. A difference was claimed only if its 95\% interval excluded 0. Each test set was opened only once, after the panel sizes and all pipeline settings had been fixed, and the time of opening was logged.

\subsection{Robustness Checks}
\label{sec:robust-methods}
Four checks were fixed before they were run.

\emph{Realized depth.} Unrestricted forests (100 trees, balanced class weights) were fitted to the candidates of every training fold of both tasks, and the depth of every tree was recorded.

\emph{Importance measure.} On pan-cancer fold 0, forests of 300 trees at depth 5 and at unlimited depth (seeds 0--2) were fitted to the 3,000 candidates. Genes were ranked by impurity importance and by permutation importance \cite{breiman2001,altmann2010} (increase in validation log loss, two permutations per gene), and the depth gap was computed for each; in this check it was the mean seed agreement at the two depths minus their cross-seed overlap. The depth effect was taken as independent of the importance measure if the permutation-importance gap was at least 0.05.

\emph{Replication.} The second prepared CV repeat (new folds, pre-filter refitted in each training fold) was run at depths 5, 8, 10, 15 and unlimited, seeds 0--2 (75 runs). The main effect counted as replicated if the depth-5 gap was at least 0.05 and positive in all five folds; the observation that depth 10 is least reproducible was kept only if depth 10 again had the lowest seed agreement among depths 8, 10 and 15.

\emph{Sample size versus classes.} In every pan-cancer training fold, we drew stratified subsamples of 637, 1,600 and 3,300 samples and used the full fold (all 33 classes), and subsamples of 637 samples restricted to 5, 10, 20 or 33 randomly drawn cancer types (three draws each); the pre-filter was refitted on each subsample and the realized depth of 100 unrestricted trees was recorded. Boruta was then run on the 637-sample, 33-class subsample at depths 5, 8, 10 and unlimited (seeds 0--2, 60 runs); depths 15 and 20 were added after this first run (30 runs) to locate where the gap vanishes. Sample size was taken as the main driver if the depth-8 gap for the top 200 genes was at most 0.02, as in PAM50 (637 training samples), and the number or kind of classes if it was at least 0.05.

\subsection{Software and Computing}
Python 3.10.12, NumPy 2.2.6 \cite{harris2020}, pandas 2.3.3, SciPy 1.15.3 \cite{virtanen2020}, scikit-learn 1.7.2 \cite{pedregosa2011}, LightGBM 4.7.0 \cite{ke2017}, joblib 1.5.3, Matplotlib 3.10.8 \cite{hunter2007} and, for the GDC download, requests 2.32.5. Exact versions of all packages are listed in the code repository. Runs were executed on a 47-core server with 500 GB RAM, 15 runs in parallel with 3 threads each. A pan-cancer Boruta run at unlimited depth took about 84 minutes per training fold (300 trees, 50 iterations); a PAM50 run took about 3 minutes.

\section{Results}
\label{sec:results}
Tree depth shifts Boruta's selection far beyond seed noise in pan-cancer data and barely at all in PAM50 subtyping; the difference follows how deep unrestricted trees grow. The quality filter removes quality-associated genes in both tasks without changing test accuracy.

\begin{figure*}[!t]
\centering
\includegraphics[width=\textwidth]{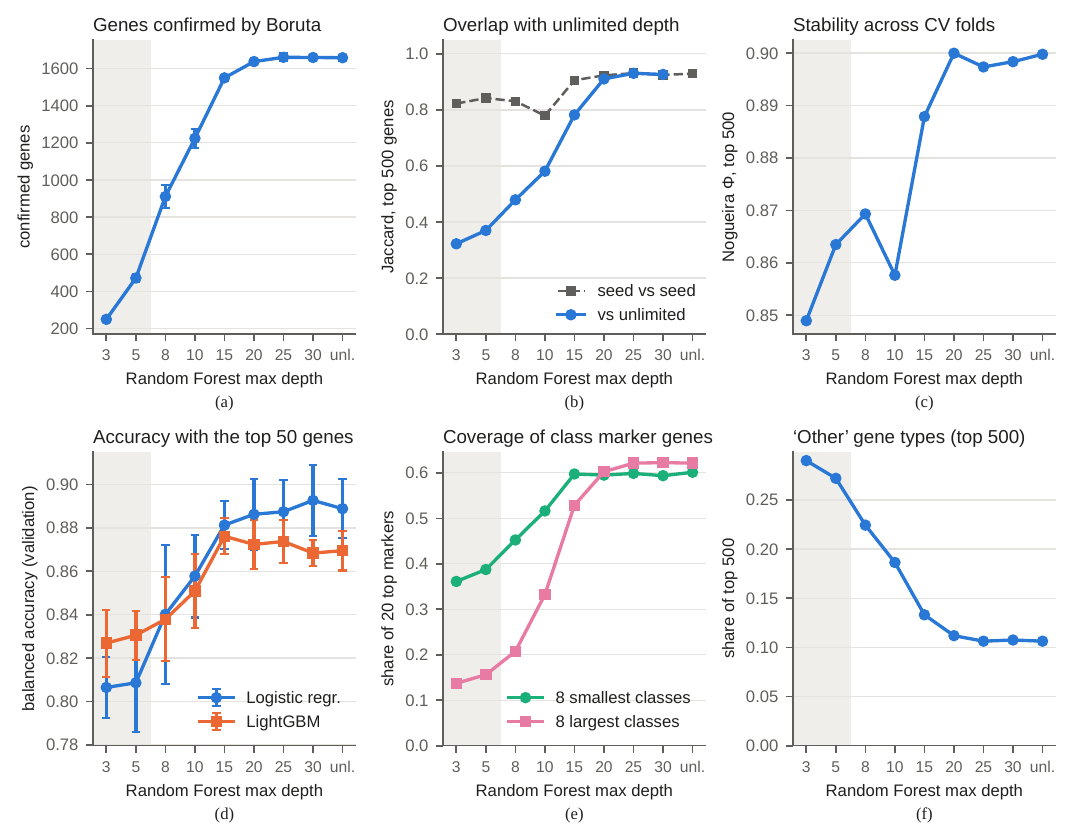}
\caption{Effect of Random Forest maximum depth on Boruta feature selection in TCGA pan-cancer data (33 classes; 5-fold cross-validation, 3 seeds per depth). Shaded: depth range recommended in the BorutaPy documentation (3--7); ``unl.'' = unlimited depth. (a) Genes confirmed by Boruta (mean $\pm$ SD over folds and seeds). (b) Jaccard overlap of the top 500 genes: blue, between each depth and unlimited depth, compared across different seeds; dashed, between two seeds at the same depth. (c) Stability of the top 500 genes across folds (Nogueira index). (d) Validation balanced accuracy of logistic regression and LightGBM trained on the top 50 genes (mean $\pm$ SD over folds); differences between depths are consistent within folds (Section~\ref{sec:res-acc}). (e) Share of each class's 20 strongest one-vs-rest marker genes found among the top 500, averaged over the 8 smallest and the 8 largest classes. (f) Share of non-coding ``other'' gene types, mostly processed pseudogenes, among the top 500.}\label{fig:depth}
\end{figure*}

\subsection{Depth Changes What Boruta Selects in Pan-Cancer Data}
In the pan-cancer task, tree depth changed the selected genes far more than the random seed did, and the pre-declared decision rule was met in the first phase (depths 5, 15 and unlimited, seeds 0--4): the top-500 overlap between depths (0.528) was 0.364 below the overlap between seeds (0.892), and lower in all five folds (differences 0.347--0.377). Depth-5 trees confirmed 472 genes against 1,659 at unlimited depth, and depth 3 only 249 (Fig.~\ref{fig:depth}(a)).

\begin{figure*}[!t]
\centering
\includegraphics[width=\textwidth]{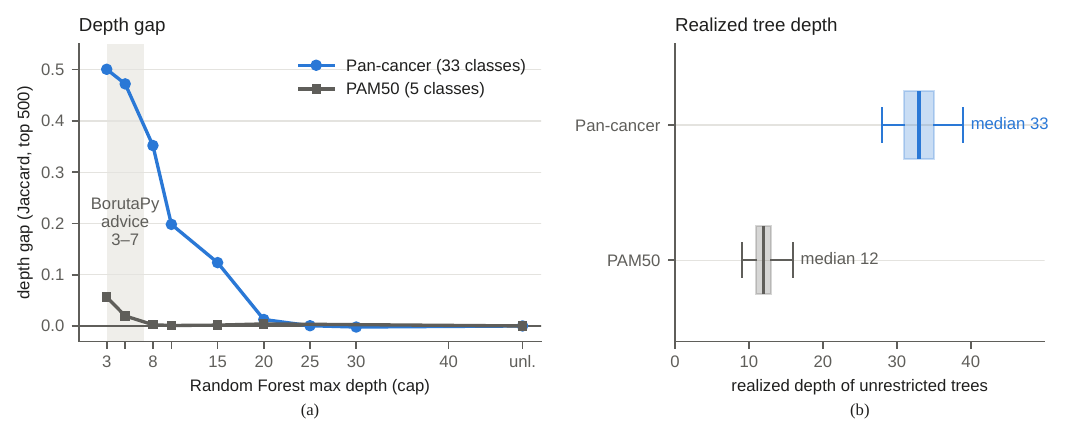}
\caption{The depth effect follows the depth that unrestricted trees reach. (a) Depth gap, defined as the seed-to-seed overlap of the top 500 genes minus their overlap with unlimited-depth selections from different seeds, in the pan-cancer and PAM50 tasks; 0 means changing the depth alters the selection no more than changing the seed. Shaded: BorutaPy-recommended range. (b) Realized depth of unrestricted Random Forest trees (100 trees per training fold, 5 folds), on the same depth scale as (a); boxes show quartiles, whiskers the 5th--95th percentiles.}\label{fig:gap}
\end{figure*}

Measured against seed noise, the depth gap closes steadily between depth 8 and depth 20 (Fig.~\ref{fig:depth}(b)). It is 0.50 at depth 3, 0.47 at depth 5 and 0.35 at depth 8, falls to 0.20 at depth 10 and 0.12 at depth 15, and is 0.01 at depth 20 and 0.00 from depth 25. Across all five training folds, unrestricted trees reached median depth 33 (5th--95th percentile 28--39; Fig.~\ref{fig:gap}). A cap of 20 therefore cuts almost every tree, yet changes the selection no more than a new seed does: splits below depth 20 add little to the importance ranking.

Every depth was reproducible across folds ($\Phi$ 0.85--0.90), so depth causes a systematic shift rather than instability, although shallow selections were somewhat less stable ($\Phi$ 0.849 at depth 3, 0.863 at depth 5, 0.900 at unlimited depth) and agreed less between seeds (0.82--0.84 vs 0.93). Depth 10 was the least reproducible setting: runs with different seeds agreed least (top-500 Jaccard 0.78) and stability across folds was low ($\Phi$ 0.858), lying in the middle of the transition; this replicated in a second CV repeat (Section~\ref{sec:res-robust}). Averaging the hit rates of depths 5, 15 and unlimited did not stabilize the selection either ($\Phi$ 0.861, below unlimited depth alone).

\subsection{Shallow Trees Give Worse Small Panels}
\label{sec:res-acc}
With 50 genes, balanced accuracy rose from 0.806 (depth 3) and 0.809 (depth 5) to 0.889 at unlimited depth with logistic regression, and from 0.827 to 0.870 with LightGBM (Fig.~\ref{fig:depth}(d)). Paired by fold, depth 5 was worse than unlimited depth at $K = 50$ by 0.080 (logistic regression) and 0.039 (LightGBM), with no fold favoring shallow trees. The gap shrank with panel size (LightGBM: $-0.018$ at $K = 100$, $-0.009$ at 500, $-0.003$ at 1,000), and at $K = 1{,}000$ every depth was within 0.927--0.943. For reference, at $K = 50$ the ANOVA F ranking reached 0.772 (logistic regression) and 0.806 (LightGBM), and random candidates 0.849 and 0.819.

\begin{table*}[!t]
\caption{Pan-Cancer Selections by Random Forest Depth (5 Folds, Seeds 0--2; Top 500 Genes Unless Stated)}
\label{tab:props}
\centering
\setlength{\tabcolsep}{5pt}
\resizebox{\textwidth}{!}{%
\begin{tabular}{lccccccccc}
\toprule
Depth & 3 & 5 & 8 & 10 & 15 & 20 & 25 & 30 & Unlimited \\
\midrule
Confirmed genes & 249 & 472 & 911 & 1,224 & 1,550 & 1,638 & 1,661 & 1,660 & 1,659 \\
Seed-to-seed Jaccard & 0.823 & 0.842 & 0.831 & 0.779 & 0.906 & 0.923 & 0.931 & 0.924 & 0.930 \\
Jaccard vs unlimited (cross-seed) & 0.322 & 0.370 & 0.479 & 0.581 & 0.782 & 0.911 & 0.931 & 0.926 & -- \\
Nogueira $\Phi$ & 0.849 & 0.863 & 0.869 & 0.858 & 0.888 & 0.900 & 0.897 & 0.898 & 0.900 \\
Balanced accuracy, LR, $K = 50$ & 0.806 & 0.809 & 0.840 & 0.858 & 0.881 & 0.886 & 0.888 & 0.893 & 0.889 \\
Balanced accuracy, LightGBM, $K = 50$ & 0.827 & 0.831 & 0.838 & 0.851 & 0.876 & 0.872 & 0.874 & 0.868 & 0.870 \\
``Other'' gene-type share & 0.29 & 0.27 & 0.22 & 0.19 & 0.13 & 0.11 & 0.11 & 0.11 & 0.11 \\
Marker coverage, 8 largest classes & 0.14 & 0.16 & 0.21 & 0.33 & 0.53 & 0.60 & 0.62 & 0.62 & 0.62 \\
Quality-associated share & 0.20 & 0.15 & 0.10 & 0.08 & 0.06 & 0.05 & 0.05 & 0.05 & 0.05 \\
\bottomrule
\multicolumn{10}{p{0.95\textwidth}}{\footnotesize Balanced accuracy on validation folds. Marker coverage: share of each class's 20 strongest one-vs-rest markers in the top 500. Quality-associated share: genes with $|r_q| > 0.3$ among the top 500 (Fig.~\ref{fig:quality}(b)).}
\end{tabular}}
\end{table*}

\subsection{What Shallow Trees Select}
\label{sec:res-props}
Gene properties changed smoothly with depth (Table~\ref{tab:props}). Shallow trees favored genes with strong univariate signal (median ANOVA F percentile 83 at depth 3, 73 at unlimited depth) and non-coding ``other'' gene types (share 0.29 vs 0.11). They missed most markers of large classes: coverage of the 20 strongest one-vs-rest markers of the eight largest classes was 0.14 at depth 3 and 0.62 at unlimited depth. Markers of small classes were recovered by depth 15 (about 0.60); those of large classes only by depth 20. Across five folds, 175 genes were specific to depth 5 (74 ``other'') and 195 to unlimited depth (150 mRNA), around a shared core of 238.

\subsection{In PAM50 Subtyping, Unrestricted Trees Are Shallow and Depth Matters Little}
With 637 training samples and five classes, unrestricted trees reached median depth 12 (range 8--20), and the depth effect was small. Only depth 3 departed clearly from unlimited depth: 264 confirmed genes against 386, a top-200 depth gap of 0.087, and fewer PAM50 genes in the top 200 (27.6 vs 30.2). At depth 5 the gap was 0.005 (top 200) and 0.020 (top 500); from depth 8 it was 0.00 (Fig.~\ref{fig:gap}). Validation accuracy showed no depth effect: depth 5 minus unlimited ranged from $-0.021$ to $+0.018$ across $K$ and classifiers, with mixed signs. In both tasks the gap vanished once the cap reached 61--76\% of the median depth of unrestricted trees.

Boruta clearly beat the ANOVA F ranking for small PAM50 panels: with 20 genes, 0.82 vs 0.71 (LightGBM) and 0.86 vs 0.77 (logistic regression). From $K = 100$ the two were equal.

\subsection{A Third of Candidates Are Quality-Associated}
\begin{figure*}[!t]
\centering
\includegraphics[width=\textwidth]{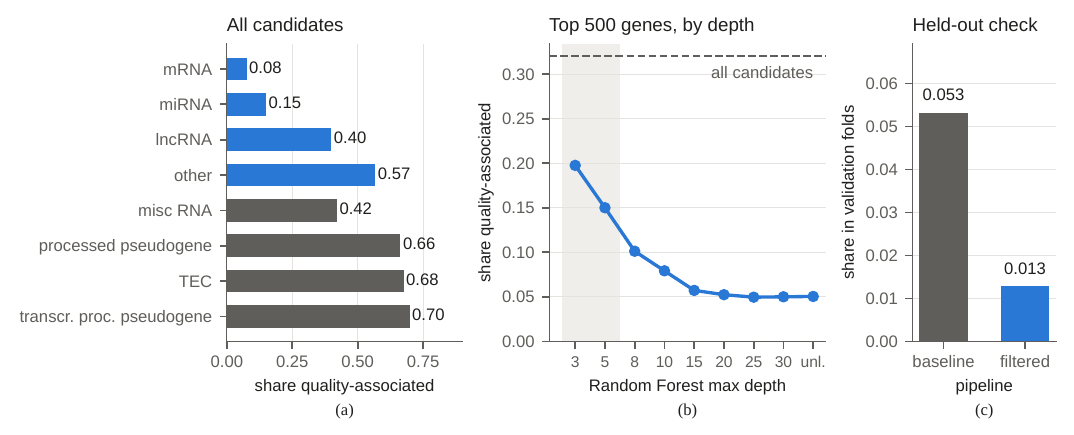}
\caption{Candidate genes associated with sample quality in the pan-cancer task. A gene is quality-associated when its within-class correlation with $\log_{10}$ of the number of detected genes exceeds 0.3 in absolute value (training fold). (a) Share of quality-associated candidates by RNA category (blue) and by the most affected gene types (gray), over 5 folds. (b) Share among the top 500 Boruta genes at each depth; dashed, share among all 3,000 candidates. (c) Share among the top 500 genes re-measured on validation samples, without and with the within-class quality filter.}\label{fig:quality}
\end{figure*}

In the pan-cancer task, 32\% of the 3,000 candidates (29--33\% per fold) had $|r_q| > 0.3$ (Fig.~\ref{fig:quality}(a)). The share depended strongly on RNA category: 8\% of mRNA candidates, 15\% of miRNA genes, 40\% of lncRNAs and 57\% of ``other'' genes, highest for transcribed processed pseudogenes (70\%), TEC (68\%) and processed pseudogenes (66\%). Category quotas therefore draw quality-associated genes into the candidate pool. Deep Boruta rejected most of them (5\% of the unlimited-depth top 500), while shallow trees kept three times as many (15\% at depth 5; Fig.~\ref{fig:quality}(b)).

Processed pseudogenes might simply mirror their parent genes through misassigned reads. They did not: within cancer type, their median correlation with the parent gene was 0.03 among all candidates.

\subsection{The Quality Filter in Cross-Validation}
The filter removed about 14,500 of 48,131 expressed genes per fold and replaced half the candidates (Jaccard with baseline 0.50--0.53), while keeping category shares. The quality-associated share of selected genes, measured again on validation samples, fell from 5.3\% to 1.3\% (Fig.~\ref{fig:quality}(c)). Selected panels held more lncRNAs (top 500: 0.216 vs 0.190). Cross-validated balanced accuracy rose slightly (LightGBM $+0.020$ at $K = 50$, higher in 5 of 5 folds; logistic regression $+0.002$ to $+0.005$). Of the two pre-declared outcomes, the first was met (the lower quality-associated share above); the second was not, because the accuracy gap between high- and low-quality validation samples did not shrink. Thresholds 0.2 and 0.4 gave similar results.

\subsection{Computational Cost}
Shallow trees are faster, but the saving buys worse panels in the pan-cancer task. With 300 trees and 50 iterations on about 6,700 training samples and 3,000 candidates, one Boruta run took 22.8 minutes at depth 5, 68.0 at depth 15 and 83.8 at unlimited depth (3.7 times longer than depth 5), on 3 CPU threads per run. In exchange, unlimited depth raised 50-gene balanced accuracy by 8.0 points (logistic regression) and 3.9 points (LightGBM). In PAM50 data the difference is small in both cost and outcome: 2.2--2.3 minutes at depth 5 against 3.0--3.1 at unlimited depth on 797 samples. The quality filter itself added no measurable time to pre-filtering, but its pan-cancer Boruta runs took slightly longer (87--93 minutes), because more genes stayed active. The full pan-cancer depth grid (165 runs) took 485 CPU-hours, and all 400 Boruta runs in this study 840 CPU-hours.

\subsection{Locked Test Sets}
\label{sec:res-test}
\begin{figure*}[!t]
\centering
\includegraphics[width=\textwidth]{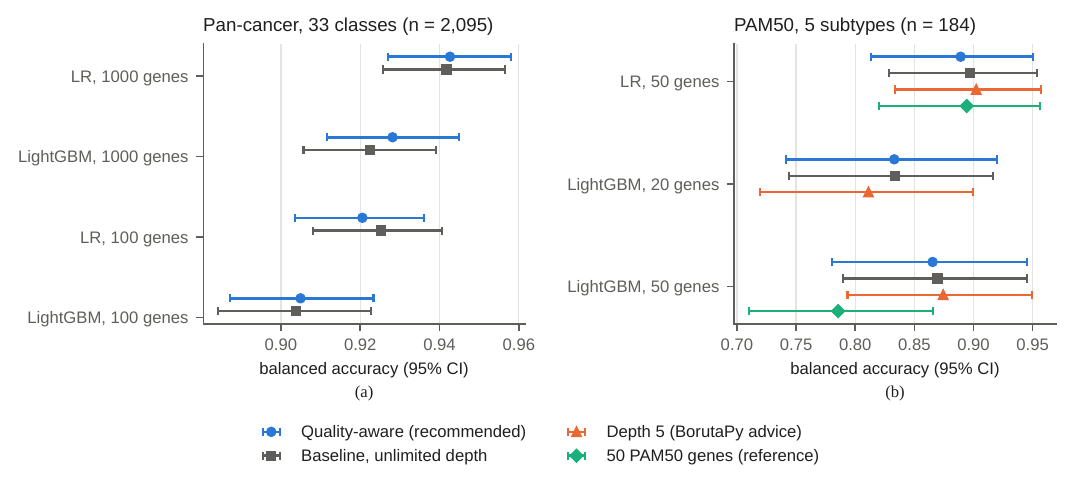}
\caption{Balanced accuracy on the locked test sets, with 95\% confidence intervals from 2,000 bootstrap resamples drawn within class. (a) Pan-cancer task (2,095 samples, 33 classes), panels of 1,000 genes (chosen by the one-standard-error rule) and 100 genes. (b) PAM50 task (184 samples, 5 subtypes), panels of 20 or 50 genes; the 50 PAM50 genes are an upper reference, as the labels were derived from them.}\label{fig:test}
\end{figure*}

On the test sets, the quality filter, and in PAM50 the depth cap, made no difference to accuracy (Table~\ref{tab:test}, Fig.~\ref{fig:test}). The one-standard-error rule chose $K = 1{,}000$ for both classifiers in the pan-cancer task, the largest size tested, and $K = 50$ (logistic regression) and $K = 20$ (LightGBM) in PAM50.

\begin{table*}[!t]
\caption{Balanced Accuracy on the Locked Test Sets}
\label{tab:test}
\centering
\setlength{\tabcolsep}{5pt}
\resizebox{\textwidth}{!}{%
\begin{tabular}{lccccc}
\toprule
Task, classifier, $K$ & Quality-aware (95\% CI) & Baseline & Depth 5 & PAM50 genes & Quality-aware minus baseline (95\% CI) \\
\midrule
Pan-cancer, LR, 1,000 & 0.943 (0.927--0.958) & 0.942 & -- & -- & $+0.001$ ($-0.006$ to 0.007) \\
Pan-cancer, LightGBM, 1,000 & 0.928 (0.912--0.945) & 0.923 & -- & -- & $+0.006$ ($-0.001$ to 0.013) \\
Pan-cancer, LR, 100 & 0.921 (0.903--0.936) & 0.925 & -- & -- & $-0.005$ ($-0.013$ to 0.002) \\
Pan-cancer, LightGBM, 100 & 0.905 (0.887--0.923) & 0.904 & -- & -- & $+0.001$ ($-0.012$ to 0.013) \\
PAM50, LR, 50 & 0.889 (0.814--0.951) & 0.897 & 0.902 & 0.894 & $-0.008$ ($-0.044$ to 0.017) \\
PAM50, LightGBM, 20 & 0.833 (0.741--0.920) & 0.834 & 0.811 & -- & $-0.001$ ($-0.039$ to 0.031) \\
PAM50, LightGBM, 50 & 0.865 (0.780--0.945) & 0.870 & 0.874 & 0.786 & $-0.004$ ($-0.053$ to 0.036) \\
\bottomrule
\multicolumn{6}{p{0.95\textwidth}}{\footnotesize Pan-cancer: 2,095 samples, 33 classes; PAM50: 184 samples, 5 classes. Rows are the pre-declared panels: the size chosen by the one-standard-error rule for each classifier and the compact panel ($K = 100$ pan-cancer, $K = 50$ PAM50). For logistic regression in PAM50 both are $K = 50$, so it has one row. Intervals from 2,000 bootstrap resamples drawn within class. LR: logistic regression; CI: confidence interval.}
\end{tabular}}
\end{table*}

In the pan-cancer task the final 1,000-gene panel reached 0.943 balanced accuracy and 0.959 accuracy with logistic regression; the 100-gene panel reached 0.921. In PAM50, of 11 paired comparisons, one interval just excluded 0 (LightGBM, $K = 20$, quality-aware minus depth 5: $+0.022$, 0.003--0.044; McNemar $p = 0.11$). It was not seen at $K = 50$ or with logistic regression, so we do not count it as a depth effect. A 50-gene Boruta panel containing 14 PAM50 genes matched the 50 PAM50 genes with logistic regression ($-0.005$, $-0.028$ to 0.019), although the labels were derived from those genes. Recall for Normal-like (7 test samples) ranged from 0.29 to 0.86 across panels.

\subsection{Robustness Checks}
\label{sec:res-robust}
The depth effect holds under permutation importance and replicates almost exactly in a second CV repeat; all three pre-declared rules (importance measure, replication, and sample size versus classes) were met.

\emph{Realized depth.} Over all folds, unrestricted trees reached median depth 33 (5th--95th percentile 28--39, range 25--44) in the pan-cancer task and 12 (9--16, range 8--20) in PAM50 subtyping. The gap vanished from depth 20--25 and from depth 8, respectively: 61--76\% and 67\% of the median.

\emph{Importance measure.} Ranking genes by permutation importance gave a depth gap of 0.223 (seed agreement 0.40 at depth 5 and 0.69 at unlimited depth; depth 5 vs unlimited 0.32), close to the gap under impurity importance (0.261). The effect is therefore not an artifact of impurity importance. The two importance rankings agreed more in deep forests (top-500 Jaccard 0.66 vs 0.55), and deep forests also classified better with all 3,000 candidates (validation balanced accuracy 0.889 vs 0.778).

\emph{Replication.} With new folds and refitted pre-filters, every measure matched the first repeat (Table~\ref{tab:repeat}). The depth-5 gap was 0.468 in both repeats (smallest fold 0.448), and depth 10 again had the lowest seed agreement (0.784, against 0.831 at depth 8 and 0.909 at depth 15). Confirmed gene counts differed by less than 1\% between repeats.

\begin{table}[!t]
\caption{Pan-Cancer Selections in Two Independent CV Repeats}
\label{tab:repeat}
\centering
\setlength{\tabcolsep}{3pt}
\begin{tabular}{lcccc}
\toprule
Depth & Depth gap & Seed Jaccard & Nogueira $\Phi$ & Confirmed \\
\midrule
5 & 0.468 / 0.468 & 0.838 / 0.839 & 0.863 / 0.872 & 472 / 469 \\
8 & 0.353 / 0.352 & 0.831 / 0.831 & 0.869 / 0.877 & 911 / 920 \\
10 & 0.199 / 0.203 & 0.779 / 0.784 & 0.858 / 0.859 & 1,224 / 1,227 \\
15 & 0.121 / 0.130 & 0.904 / 0.909 & 0.888 / 0.893 & 1,550 / 1,554 \\
Unl. & 0.000 / 0.000 & 0.927 / 0.932 & 0.900 / 0.907 & 1,659 / 1,670 \\
\bottomrule
\multicolumn{5}{p{0.95\columnwidth}}{\footnotesize Values: repeat 0 / repeat 1 (5 folds $\times$ 3 seeds each; top-500 Jaccard; depth gap measured across different seeds). Unl.: unlimited depth.}
\end{tabular}
\end{table}

\begin{figure*}[!t]
\centering
\includegraphics[width=\textwidth]{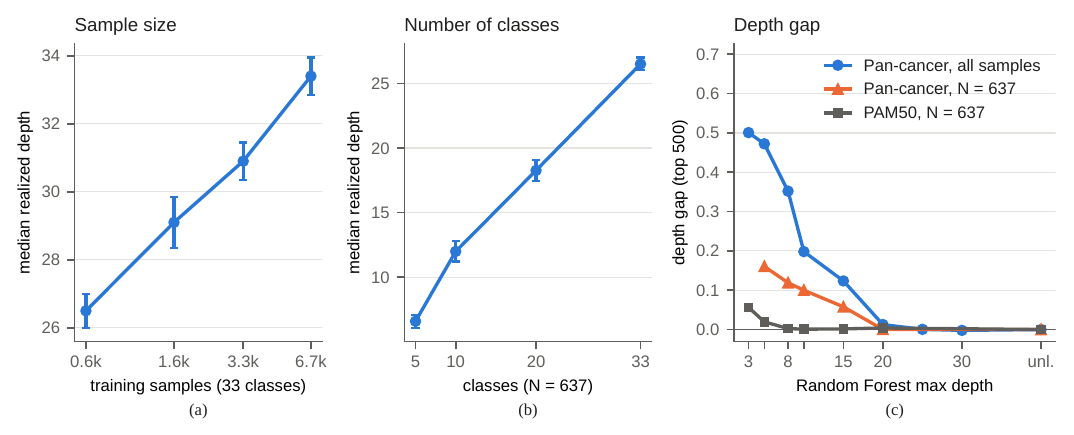}
\caption{Realized depth and the depth effect versus sample size and number of classes (pan-cancer subsamples). (a) Median realized depth of unrestricted trees (100 trees per training fold) against training-set size, all 33 classes; mean $\pm$ SD over 5 folds. (b) Median realized depth against the number of randomly drawn cancer types at 637 training samples (3 draws per fold). (c) Depth gap of the top 500 genes for the full pan-cancer data, for a 637-sample subsample with all 33 classes (depths 5--20 and unlimited), and for PAM50 (637 training samples, 5 subtypes). At 637 samples fewer than 500 genes were confirmed, so the top 500 also includes tentative genes ranked by hit rate.}\label{fig:size}
\end{figure*}

\emph{Sample size versus classes.} The two tasks differ in both sample size and number of classes, so we separated the two by subsampling the pan-cancer training folds (Fig.~\ref{fig:size}). With all 33 classes, a tenfold increase in samples raised the median realized depth only from 26.5 (637 samples) to 33.4 (about 6,700). At a fixed 637 samples, the number of classes raised it from 6.6 (5 randomly drawn cancer types) to 26.5 (33 types). With 637 samples and all 33 classes, Boruta's depth effect stayed large: the depth-8 gap was 0.292 for the top 200 genes (the set fixed in the rule; Fig.~\ref{fig:size}(c) shows the top 500), against 0.001 in PAM50 with the same number of training samples. The gap fell to 0.084 at depth 15 and 0.016 at depth 20, that is, it vanished at 75\% of the median realized depth of 26.5, in line with the two full tasks. By the pre-declared rule, the number and kind of classes, not sample size, drive the effect. Five randomly drawn cancer types gave shallower trees than the five PAM50 subtypes (median 6.6 against 12), which suggests that classes that are harder to separate also need deeper trees.

\section{Discussion}
\label{sec:discussion}

\subsection{How Depth Changes Boruta's Selection}
\label{sec:disc-mech}
Tree depth decides which genes can earn importance in Boruta. A tree capped at depth 5 has at most 31 internal splits, and with 33 classes it spends them on the largest differences: genes with strong univariate signal that separate broad groups of cancer types. Genes that distinguish large classes that resemble each other rarely get a split, so they rarely beat the shadow features and are seldom confirmed. This explains the three properties of shallow selections: a bias toward high ANOVA F genes, the loss of large-class markers (coverage 0.14 at depth 3 against 0.62 unlimited) and the weaker 50-gene panels (Sections~\ref{sec:res-acc} and \ref{sec:res-props}). Depth 10 was the least reproducible setting in both CV repeats: two runs with different seeds agreed least there (top-500 Jaccard 0.78). At that depth, about half of the large-class markers are reached (coverage 0.33 against 0.62 unlimited), and which half is reached likely depends on the random draws, so the transition region is where Boruta is most sensitive to randomness.

\subsection{When Depth Matters}
How much a depth cap matters depends on how deep unrestricted trees grow in a given problem. The effect was large in pan-cancer classification, with 33 classes and about 6,700 training samples per fold, where unrestricted trees reach median depth 33, and almost absent in PAM50 subtyping, with 5 classes and 637 training samples, where they reach median depth 12. Subsampling showed that the number of classes, not sample size, set this difference: with all 33 classes, even 637 training samples gave trees of median depth 26.5 and a large depth effect, whereas tenfold more samples added only about seven levels. In all three settings (pan-cancer, PAM50, and the 637-sample pan-cancer subsample) the depth gap vanished once the cap reached 61--76\% of the median realized depth; the coarse depth grid means the true point may lie somewhat lower. With three settings this is an observation rather than a law, but the practical advice needs no threshold: leaving depth uncapped cost no accuracy or stability in either task, and fitting one unrestricted forest shows the realized depth before any cap is chosen. The extra cost is moderate: an unlimited-depth run took 3.7 times longer than a depth-5 run in the pan-cancer task (33 classes, about 6,700 training samples) and about 1.3 times longer in PAM50 (5 classes, 797 samples).

The depth effect also survived the two checks a reader is most likely to ask for: a different importance measure and a different set of CV folds. Ranking genes by permutation importance instead of impurity importance gave nearly the same depth gap (0.223 against 0.261), so the effect is not an artifact of impurity importance. A second CV repeat with new folds and refitted pre-filters reproduced every value within 0.01.

\subsection{Relation to Prior Work and to the BorutaPy Default}
Our PAM50 result shows that where unrestricted trees are shallow, a depth cap matters little. Comparisons of Random Forest selection methods have mostly used smaller data sets with few classes \cite{kursa2014,degenhardt2019}, which may explain why depth has received little attention. The pan-cancer result shows that this small effect does not carry over to larger, multi-class problems. Kursa found that deeper Random Ferns confirmed more genes \cite{kursa2014}, matching our confirmed counts (249 at depth 3 against 1,659 unlimited). The depth range recommended by BorutaPy lies inside the shallow regime for pan-cancer data and produced worse small panels there. Comparisons limited to deep caps (15--25) or to large panels would miss the effect, because the differences there are small: with 1,000 genes, every depth reached 0.927--0.943 balanced accuracy. The test accuracy of our 1,000-gene panel (0.943 balanced accuracy, 0.959 accuracy) lies in the range reported for TCGA pan-cancer classifiers \cite{li2017,mostavi2020}, although differences in cohorts, classes and metrics prevent a direct comparison.

\subsection{Quality-Associated Genes}
About a third of pre-filtered candidates tracked the number of detected genes within class, mostly lncRNAs and pseudogenes. This is consistent with the unwanted variation documented in TCGA \cite{molania2023}. Category-aware quotas, meant to give non-coding genes a fair share of candidates, pull these genes in: each category's quota must be filled even when its genes carry weak class signal, and quality-linked variation can raise their F statistics. The within-class filter removes them before selection without dropping whole gene types, which matters because many non-coding genes carry real signal \cite{han2014}. It left test accuracy unchanged in both tasks, so we claim cleaner panels at no cost rather than higher accuracy. The small CV gains (up to 2 points with LightGBM) did not hold on the test sets, and the filter did not make classifiers more robust to sample quality within TCGA. Detected-gene counts also reflect tumor purity \cite{aran2015} and transcriptional activity, so we use ``quality-associated'' rather than ``artifact''. Boruta with unrestricted trees was partly self-cleaning: in the baseline pipeline, 5\% of its top 500 genes were quality-associated, against 15\% at depth 5, so the filter matters most for shallow trees and for univariate selectors.

\subsection{Beyond Gene Expression}
The explanation in Section~\ref{sec:disc-mech} does not depend on biology: a capped tree has few splits, and they go to the features with the strongest marginal signal. Other selectors that threshold tree importance, including other shadow-feature methods, may therefore show the same bias. In our data the effect grew with the depth that unrestricted trees reach, which rose with the number of classes and, in one comparison, was greater for classes that are harder to separate. Many-class problems in other domains, such as text, spectral or imaging data, may therefore be affected as well. We tested only RNA-seq data with Boruta, so this is a hypothesis for future work, not a finding of this study.

\subsection{Practical Guidelines for Boruta-Based Gene Selection}
\begin{enumerate}
\item Fit one unrestricted forest on the training data and record tree depths before choosing a cap.
\item If depth must be capped for speed, keep the cap above about 75\% of the median realized depth; otherwise leave it unlimited.
\item Fix the number of trees across settings; BorutaPy's automatic tree count changes with depth.
\item Judge any setting against seed variation, comparing runs with different seeds.
\item Test candidate features for within-class association with sample-quality metrics, and filter inside training folds.
\item Keep every label-dependent step inside CV folds, and open the test set only once, after all choices are fixed.
\end{enumerate}

\subsection{Limitations}
We studied one selector, Boruta with impurity importance, and both tasks come from one consortium and one processing pipeline; an external cohort is planned for a follow-up study. Permutation importance was checked on one pan-cancer fold, outside Boruta, and the second CV repeat covered five of the nine depths. Boruta was run with 300 trees and 50 iterations, and in the pan-cancer task many genes were still tentative at the end. The 0.3 quality threshold was chosen during exploration, from quality association alone; thresholds of 0.2 and 0.4 gave similar results. The class-number comparison used randomly drawn cancer types, which are easier to separate than subtypes of one cancer; class separability was not varied directly. Some PAM50 test classes are small (7 Normal-like and 13 Her2 samples), so their recall is noisy. PAM50 labels were derived from the PAM50 genes, so the PAM50 gene panel is an upper reference, not a fair competitor.

\section{Conclusion}
\label{sec:conclusion}
This study addressed two gaps in Boruta-based gene selection. First, Random Forest tree depth is not a minor setting in multi-class RNA-seq data. In TCGA pan-cancer data, the depths recommended by BorutaPy changed which genes were selected far more than the random seed did and gave less accurate small panels. The effect vanished once the cap reached 61--76\% of the median depth that unrestricted trees grow to, a depth set mainly by the number of classes rather than by sample size, and it was small in breast cancer subtyping, where unrestricted trees are shallow. Leaving depth uncapped cost no accuracy in either task. Second, about a third of pre-filtered candidates tracked sample quality within class; a within-class filter applied before the category quotas removed them and admitted more lncRNAs without loss of test accuracy. We therefore recommend an expression filter, a within-class quality filter, category-aware quotas and Boruta with unrestricted trees, with rankings averaged over seeds. The next steps are to test these panels in independent cohorts processed with other pipelines and to compare other importance-based selectors.

\section*{Author Contributions}
Tareque Mohmud Chowdhury conceived and designed the approach, performed the experiments, analyzed the data, and wrote the manuscript. Abu Raihan Mostofa Kamal supervised the whole study and revised the manuscript. Both of the authors have read and approved the final version of the manuscript.

\section*{Data and Code Availability}
{\sloppy TCGA RNA-seq data and clinical data (recorded primary diagnosis, used only to stratify the splits) are available from the NCI Genomic Data Commons (https://portal.gdc.cancer.gov; Data Release 46.0). PAM50 labels are from the TCGA PanCancer Atlas via cBioPortal (study brca\_tcga\_pan\_can\_atlas\_2018). Sample lists, splits, gene rankings and test predictions are provided as Supplementary Data. The code is available from the corresponding author upon reasonable request.\par}

\section*{Declaration of Competing Interest}
The authors declare no known competing financial interests or personal relationships that could have appeared to influence the work reported in this paper.

\bibliographystyle{IEEEtran}
\bibliography{refs}
\end{document}